\documentclass{article}

\usepackage[final]{corl_2020} 

\title{Harnessing Distribution Ratio Estimators for Learning Agents with Quality and Diversity}

\author{
  Tanmay Gangwani\\
  Dept. of Computer Science\\
  UIUC\\
  \texttt{gangwan2@illinois.edu}\\
     \And
   Jian Peng \\
   Dept. of Computer Science \\
   UIUC \\
   \texttt{jianpeng@illinois.edu} \\
    \And
   Yuan Zhou\\
   Dept. of ISE\\
   UIUC \\
   \texttt{yuanz@illinois.edu} \\
}

\usepackage{format}

\begin{document}
\maketitle

\begin{abstract}
    Quality-Diversity (QD) is a concept from Neuroevolution with some intriguing applications to Reinforcement Learning. It facilitates learning a population of agents where each member is optimized to simultaneously accumulate high task-returns and exhibit behavioral diversity compared to other members. In this paper, we build on a recent kernel-based method for training a QD policy ensemble with Stein variational gradient descent. With kernels based on $f$-divergence between the stationary distributions of policies, we convert the problem to that of efficient estimation of the ratio of these stationary distributions. We then study various distribution ratio estimators used previously for off-policy evaluation and imitation and re-purpose them to compute the gradients for policies in an ensemble such that the resultant population is diverse and of high-quality~\footnote{Code for this paper is available at \url{https://github.com/tgangwani/QDAgents}}.
\end{abstract}



\section{Introduction}
\label{sec:intro}
\begin{figure}[t]
\centering
\captionsetup[subfigure]{justification=centering}
    \begin{subfigure}[t]{0.2\textwidth}
    \centering
        \includegraphics[scale=0.1]{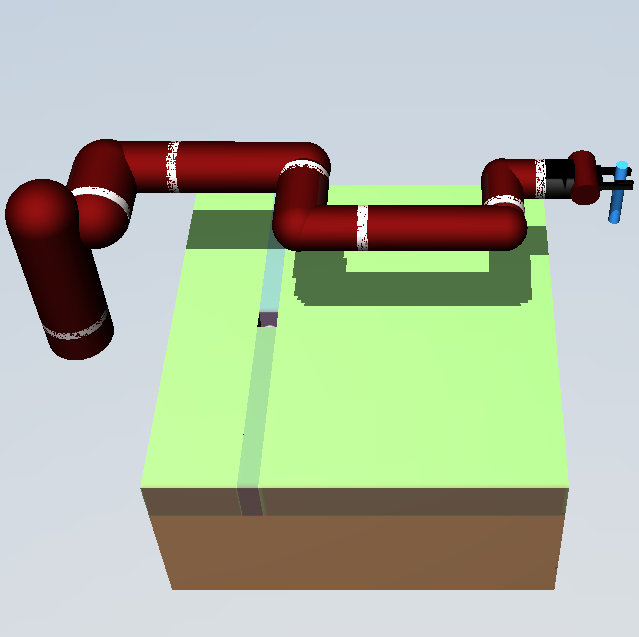}
        \caption{}
        \label{fig:sawyer_bot}
    \end{subfigure}\hfill
    \begin{subfigure}[t]{0.8\textwidth}
        \centering
        \includegraphics[scale=0.55]{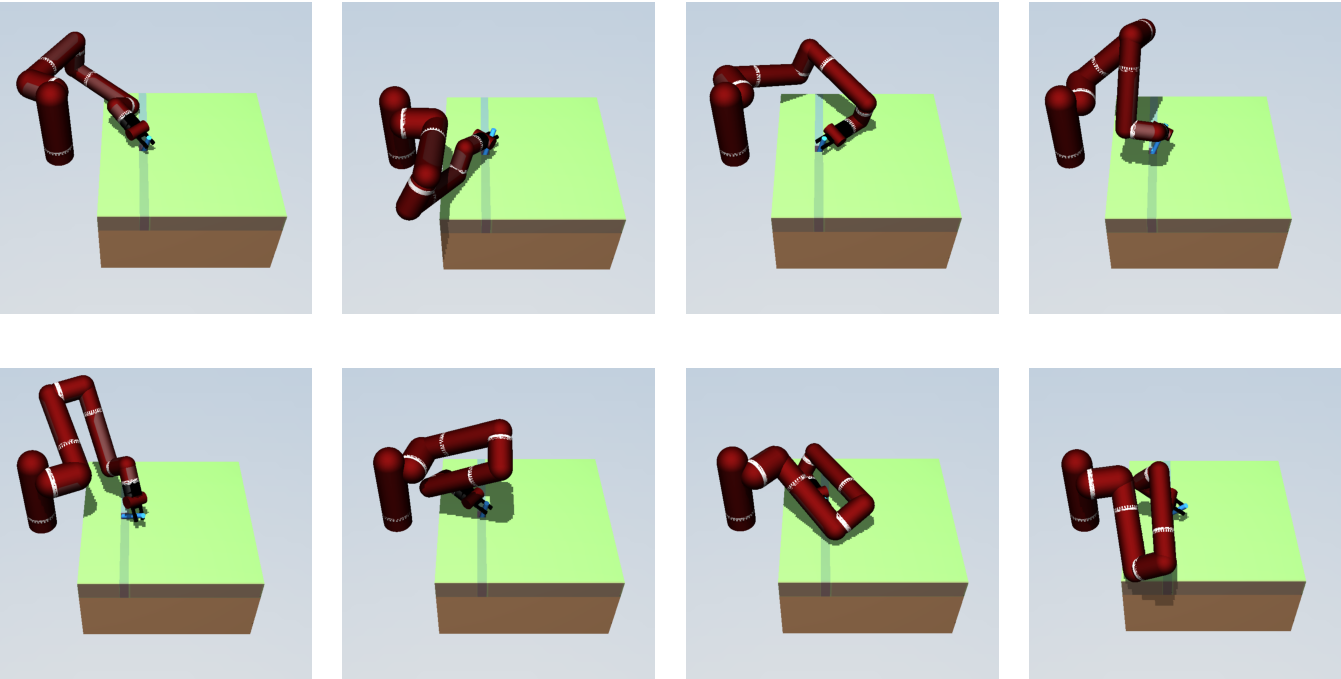}
        \caption{}
        \label{fig:sawyer_dd_js}
    \end{subfigure}\hfill
    \caption{(a) MuJoCo model of a 7 DOF arm based on the Sawyer robot, inspired by~\citet{chen2018hardware}; (b) Policies that achieve the peg-insertion task in different ways. These policies are sampled from a single ensemble trained with the algorithm QD-{\em DualDICE}-JS (explained in Section~\ref{sec:exp}).}
    \label{fig:intro_full}
    \vspace{-3mm}
\end{figure}

The goal in Reinforcement Learning (RL) is to learn agents that maximize long-term environmental rewards. Deep RL, which uses deep neural networks as function approximators for the policy and value-functions, has achieved outstanding results on a wide variety of sequential decision making problems, with the barometer of success usually being the total returns accumulated by the final policy. Due to the intrinsic nature of direct reward maximization, seldom is the focus on how the {\em behavioral characteristics} of the trained agent compare with the other possible behaviors in the solution space. For instance, consider the robotic manipulator arm in Figure~\ref{fig:sawyer_bot} and the peg-insertion task. Though the task description is simple, for a sufficiently flexible arm, there are numerous ways (positions of the joints and the end-effector) to insert the peg in the hole (Figure~\ref{fig:sawyer_dd_js}). For reasons argued below, it is beneficial to learn these varied behaviors rather than aiming for the single most efficient solution dictated by the reward function. Quality-Diversity (QD) algorithms~\citep{pugh2016quality, cully2017quality} are
prominent in the Neuroevolution literature as a means to generate many diverse behavioral {\em niches}, while ensuring that each niche is populated with individuals of the highest possible quality for that niche. When applied to RL, QD offers a principled approach for learning policies that are diverse, yet achieve high returns~\citep{mouret2015illuminating,conti2017improving}.

Prior works have examined the benefits of uncovering diversity in how the task can be solved~\citep{hong2018diversity,eysenbach2018diversity,liu2017stein}. In these, an explicit diversity-maximization objective is incorporated into the RL algorithm to facilitate the learning of diverse {\em skills}. There are several important benefits of training a population of agents with diverse skills. Firstly, this is an efficient exploration strategy in sparse-reward environments as the agents can collectively achieve much wider coverage of the state-space, while reducing the susceptibility of RL to local optimal solutions caused by deceptive rewards~\citep{conti2017improving, gangwani2018learning}. Secondly, the acquired skills could be leveraged for accelerated learning in downstream tasks, for example, by composing the skills to solve long-horizon tasks via hierarchical RL~\citep{florensa2017stochastic,eysenbach2018diversity}. Diversity also helps in the transfer learning of RL policies across environments that may have discrepancies such as system dynamics mismatch. Having a repertoire of skills is useful when knowledege transfer is done to a target environment that has constraints on the set of feasible behaviors~\citep{cully2015robots}.

A policy $\pi$ is characterized by its occupancy measure $\rho_\pi$~\citep{Puterman}, which is the stationary distribution over the state-action pairs that $\pi$ encounters when navigating the environment. Given two policies $\pi, \beta$, the ratio of their stationary distributions $\zeta = \rho_\pi / \rho_\beta$ is a well-studied quantity in RL. Estimates of the distribution ratio are useful for off-policy evaluation~\citep{precup2000eligibility, thomas2016data} (where the goal is to evaluate the performance of $\pi$ using fixed data generated from $\beta$), policy optimization~\citep{sutton2016emphatic,liu2019off} and off-policy imitation learning~\citep{kostrikov2019imitation}. In this paper, we examine the use of distribution ratio estimators for learning a diverse policy ensemble with high returns (a QD ensemble). We build on the approach introduced by~\citet{liu2017stein}. Using Stein variational gradient descent (SVGD)~\citep{liu2016stein} as the inference method, the authors construct an update rule that includes the policy-gradient on the environmental rewards (for high quality) and a kernel-induced repulsive force gradient (for high diversity). This kernel-based algorithm is naturally impacted by the choice of the kernel. We begin with generalizing the Stein variational policy gradient (SVPG) objective~\citep{liu2017stein} by using as kernels the negative exponents of an $f$-divergence between the stationary distributions of two policies, and discuss key properties such as positive-definiteness of kernels. For kernels based on the Jenson-Shannon and Symmetric Kullback-Leibler divergences, we show how the complete SVPG gradient can be recast in terms of the {\em ratio of the stationary distributions} ($\zeta$) between policies. Then, to estimate these ratios, and hence the SVPG gradient, we study three recently proposed distribution ratio estimators for off-policy evaluation and imitation learning. These are {\em DualDICE}~\citep{nachum2019dualdice}, {\em ValueDICE}~\citep{kostrikov2019imitation} and {\em GenDICE}~\citep{zhang2020gendice}. Additionally, we describe a fourth estimator based on Noise-Constrastive Estimation~\citep{gutmann2010noise}.

We perform experiments on various tasks to get a measure of the effectiveness of our proposed approach in generating diverse behaviors with high returns. We also evaluate on tasks with deceptive rewards and those which lack an external reward signal to further illuminate the benefits of QD.










\section{Preliminaries}
\label{sec:prelim}
\textbf{RL Notations.} The environment is modeled as an infinite-horizon, discrete-time Markov Decision Process (MDP), represented by the tuple ($\mathcal{S}$, $\mathcal{A}$, $\mu_0$, $r$, $p$, $\gamma$), where $\mathcal{S}$ is the state-space, $\mathcal{A}$ is the action-space, $\gamma\in[0,1)$ is the discount factor, and $\mu_0$ denotes the initial state distribution. Given an action $a_t$ sampled from a stochastic policy $\pi_\theta(a_t|s_t)$, the next state is sampled from the transition dynamics distribution, $s_{t+1} \sim p(s_{t+1}|s_t,a_t)$, and the agent receives a reward $r(s_t,a_t)$ determined by the reward function $r: \mathcal{S}\times\mathcal{A} \rightarrow \mathbb{R}$. The RL objective is to maximize the expected discounted sum of rewards, $\eta(\pi) = (1-\gamma)\mathbb{E}_{\mu_0, p, \pi} \big[ \sum_{t=0}^{\infty} \gamma^t r(s_t,a_t) \big]$.

\textbf{Distribution Ratio $(\zeta)$.} The occupancy measure~\citep{Puterman}, or the stationary discounted state-action visitation distribution of the policy $\pi$ is defined as $\rho_{\pi}(s,a) = (1-\gamma)\pi(a|s)\sum_{t=0}^{\infty} \gamma^{t} \mathbb{P}(s_t{=}s|\pi)$, where $\mathbb{P}(s_t{=}s|\pi)$ is the probability of being in state $s$ at time $t$, when starting in state $s_0 \sim \mu_0$ and using $\pi$ thereafter. The stationary distribution~\footnote{Throughout, stationary {\em discounted} distribution is shorthanded with stationary distribution for brevity} affords a convenient rewriting of the expected policy return as $\eta(\pi) = \mathbb{E}_{\rho_{\pi}} [r(s,a)]$, and the gradient is provided by the policy gradient theorem~\citep{sutton2000policy} as $\nabla_{\theta} \eta(\pi) = \mathbb{E}_{\rho_{\pi}} \big[\nabla_{\theta}\log \pi_\theta(a|s)Q^{\pi}(s,a)\big]$, where $Q^{\pi}$ is the state-action value function. For two policies $\pi_i$ and $\pi_j$, we denote the ratio of their stationary distributions by $\zeta_{ij}(s,a) = \rho_{\pi_i}(s,a)/\rho_{\pi_j}(s,a)$. This ratio is widely applicable for off-policy evaluation as it enables estimating the expected returns of $\pi_i$ using a fixed dataset $\mathcal{D}$ of transitions generated from a different behavioral policy $\pi_j$, since $\eta(\pi_i) = \mathbb{E}_{(s,a)\sim\mathcal{D}} [\zeta_{ij}(s,a)r(s,a)]$, where $\mathcal{D}$ is an empirical estimate of $\rho_{\pi_j}$.	

\section{Methods}
\label{sec:methods}
{\renewcommand{\arraystretch}{1.3} 
\begin{table}[]
\centering
\resizebox{\columnwidth}{!}{%
\begin{tabular}{cccc}
Name of $f$-divergence & Formula $D_f$(P, Q) & Generator $f(u)$ with $f(1) = 0$ & \shortstack{Is the kernel \\ $e^{-D_f(P, Q)/T}$ PD?} \\ \hline \hline
   Jenson-Shannon &  $\int_{x}\frac{p(x)}{2} \log \frac{2p(x)}{p(x) + q(x)} + \frac{q(x)}{2} \log \frac{2q(x)}{p(x) + q(x)}dx$ &  $\frac{u}{2}\log u -\frac{(1+u)}{2}\log \frac{1+u}{2}$ & Yes \\       
   Triangular Discrimination & $\int_{x}\frac{(p(x)-q(x))^2}{p(x)+q(x)}dx$ & $\frac{(u-1)^2}{u+1}$ & Yes \\
   Squared Hellinger & $\int_{x}(\sqrt{p(x)} - \sqrt{q(x)})^2dx$ & $(\sqrt{u}-1)^2$ & Yes \\
   Total Variation & $\frac{1}{2}\int_{x}|p(x)-q(x)|dx$ &  $\frac{1}{2}|u-1|$ & Yes \\
   Kullback-Leibler & $\int_{x}p(x)\log\frac{p(x)}{q(x)}dx$ & $- \log u$ & No \\
   Reverse Kullback-Leibler & $\int_{x}q(x)\log\frac{q(x)}{p(x)}dx$ & $u \log u$ & No \\
   Symmetric Kullback-Leibler & $\int_{x}p(x)\log\frac{p(x)}{q(x)} + q(x)\log\frac{q(x)}{p(x)}dx$ & $(u-1) \log u$ & No
\end{tabular}%
}
\vspace{2mm}
\caption{$f$-divergences and positive-definiteness of the negative exponential kernels.}
\label{tab:divergences}
\vspace{-4mm}
\end{table}}

\subsection{QD objective and its solution based on variational inference}
QD when applied to policy search entails learning multiple policies that all accumulate high environmental rewards during an episode, but the agents accomplish this using diversified strategies, such as navigating dissimilar regions of the state-action space. 
Formally, policy search with QD could be defined as learning a {\em distribution} over the policy parameters ($\theta$) that maximizes the RL-objective in expectation, while maintaining a high-entropy ($\mathcal{H}$) in the parameter-space:
\begin{equation}\label{eq:param_diversity}
    \max_q \mathbb{E}_{\theta\sim q}[\eta(\theta)] + \mathcal{H}(q) \:; \quad \quad \mathcal{H}(q) = \mathbb{E}_{\theta \sim q} [-\log q(\theta)]
\end{equation}
Solving the objective in Equation~\ref{eq:param_diversity} analytically yields the following energy-based optimal parameter distribution: $q^{*}(\theta) = \exp(\eta(\theta)) / Z_{q^{*}}$, where $Z_{q^{*}}$ is the normalization constant. Let $p(\theta)$ define a trainable distribution over the policy parameters that we seek to optimize to be close ({\em w.r.t.} the KL-divergence) to the target distribution $q^{*}$. Representing $p(\theta)$ with a mixture of delta distributions, the variational objective is:
\[
\min_p D_{\text{KL}} \big[p \:||\: \exp(\eta(\theta)) / Z_{q^{*}}\big]\:; \quad \quad p(\theta) = \frac{1}{n}\sum_{i=1}^n \delta(\theta=\theta_i)
\]
Here $\{\theta_i\}^n_1$ denotes a policy ensemble with $n$ discrete policies that constitute the $p$ distribution. Stein variational gradient descent (SVGD; \citet{liu2016stein}) provides an efficient solution to obtain an approximate gradient on the $p$ distribution. Suppose we perturb each policy $\theta_i$ with $\Delta \theta_i$ such that the induced KL between $p$ and $q$ is reduced. The optimal perturbation direction, in the unit ball of a reproducing kernel Hilbert space associated with a kernel function $k$, that maximally decreases the KL is given by~\citep{liu2016stein}:
\[
\Delta \theta = \mathbb{E}_{\theta'\sim p} \big[ \nabla_{\theta'} \log q^{*}(\theta') k(\theta', \theta) + \nabla_{\theta'} k(\theta', \theta) \big]
\]
Using this result and the energy-based form of the target distribution $q^{*}$, SVPG~\citep{liu2017stein} iteratively updates the policies with the following rule to learn a policy ensemble with QD behavior:
\begin{align}\label{eq:svpg}
\theta_i \gets \theta_i + \epsilon \Delta \theta_i,
&& 
\Delta \theta_i = \frac{1}{n} \sum_{j=1}^n \big[ \underbrace{\nabla_{\theta_j} \eta(\pi_{\theta_j}) k(\theta_j, \theta_i)}_{\text{Quality-enforcing}} + \underbrace{\nabla_{\theta_j} k(\theta_j, \theta_i)}_{\text{Diversity-enforcing}} \big] 
\end{align}

\subsection{Negative exponents of $f$-divergences as kernels}
The positive definite (PD) kernel function $k$ in Equation~\ref{eq:svpg} is an algorithmic design choice. There are two considerations. It should be possible to efficiently compute $k(\theta_j, \theta_i)$ for any two policies $(\pi_{\theta_j}, \pi_{\theta_i})$ as well as its gradient {\em w.r.t.} the policy parameters; and the function should be sufficiently expressive to capture the complex interactions between policy behaviors.~\citet{liu2017stein} employ a Gaussian RBF kernel $k(\theta_j, \theta_i)=\exp (-{\Vert \theta_j - \theta_i \Vert}_2^2 / h)$, with a dynamically adapted bandwidth $h$.~\citet{gangwani2018learning} suggest replacing the Euclidean distance in the parameter space with a statistical distance in the stationary distribution space, and use $k(\theta_j, \theta_i) = \exp (-D_{\text{JS}}(\rho_{\pi_{\theta_j}}, \rho_{\pi_{\theta_i}})/\mathit{T})$, where $D_{\text{JS}}$ is the Jenson-Shannon divergence and $\mathit{T}$ is the temperature. $D_{\text{JS}}$ is a member of a broader class of divergences known as Ali-Silvey distances or $f$-divergences~\citep{ali1966general}. Given two distributions with continuous densities $p(x)$ and $q(x)$ over the support $\mathcal{X}$, the $f$-divergence between them is defined as:
\[
    D_{f}(p\:||\:q) = \int_{\mathcal{X}} q(x) f(\frac{p(x)}{q(x)}) dx  
\]
where $f : \mathbb{R}_{+} \rightarrow \mathbb{R}$ is a convex, lower-semicontinuous function such that $f$(1) = 0. Different choices for the function $f$ recover the well-known divergences. Although generalizing the kernel function as $k_f(\theta_j, \theta_i) = \exp (-D_{f}(\rho_{\pi_{\theta_j}}, \rho_{\pi_{\theta_i}})/\mathit{T})$ may seem like a natural extension, for some $f$-divergences, $k_f(\theta_j, \theta_i)$ is not PD, and hence not a proper kernel from a theoretical standpoint. For instance, while $k_\text{JS}$ is PD, kernels with other divergences commonly used for policy learning (KL, Reverse-KL) are not. Table~\ref{tab:divergences} provides details on the various divergences that define PD and non-PD kernels after negative exponentiation. The first four divergences in Table~\ref{tab:divergences} are squared metrics ({\em i.e.} $\sqrt{D_f}$ is a true metric) and the proof of positive-definiteness of the corresponding kernels $k_f$ is provided in~\citet{hein2004hilbertian}. Inserting $k_f(\theta_j, \theta_i)$ in Equation~\ref{eq:svpg}, the SVPG gradient becomes:
\begin{equation}\label{eq:svpg_general}
\Delta \theta_i = \frac{1}{n} \sum_{j=1}^n \exp (-\underbrace{D_{f}(\rho_{\pi_{\theta_j}}, \rho_{\pi_{\theta_i}})}_{\text{Divergence value}} /\mathit{T}) \Big[ \underbrace{\nabla_{\theta_j} \eta(\pi_{\theta_j})}_{\text{Policy gradient}} - \frac{1}{\mathit{T}} \underbrace{\nabla_{\theta_j} D_{f}(\rho_{\pi_{\theta_j}}, \rho_{\pi_{\theta_i}})}_{\text{Divergence gradient}} \Big]
\end{equation}

This provides a general framework to evaluate the SVPG gradient for learning a QD policy ensemble. Depending on the $f$-divergence and the method for estimating its value and gradient, several approaches are possible, a few of which we will discuss. We use the shorthand notation $\rho_{i}$ for the stationary distribution of the policy $\pi_{\theta_i}$. Then $\zeta_{ij}(s,a) = \rho_{i} (s,a) / \rho_{j} (s,a)$ is the distribution ratio for two given policies, and would be the pivotal quantity in the exposition that follows. Next, we rewrite two kernels (and their gradient {\em w.r.t.} the policy parameters) in terms of $\zeta$, before elucidating several methods to estimate $\zeta$ for use in a practical RL algorithm to generate a QD policy ensemble.

\noindent
\textbf{The $k_\text{JS}$ and $k_\text{KLS}$ kernels.} While $k_\text{JS}$ is a PD kernel, $k_\text{KLS}$ is not since
$\sqrt{D_\text{KLS}}$ is not a metric as it does not satisfy the triangle inequality. Although positive-definiteness is desirable, non-PD kernels may yet achieve good performance in practice, as shown in~\citet{moreno2004kullback}, where SVM classification with a non-PD kernel leads to better accuracy than provably PD kernels. Both $k_\text{JS}$ and $k_\text{KLS}$ afford the benefit that the divergence value and gradient (in Equation~\ref{eq:svpg_general}) can be evaluated in terms of the distribution ratio $\zeta_{ij}$. Using the definitions from Table~\ref{tab:divergences}, we express $D_{\text{JS}}$ and $D_{\text{KLS}}$ as:
\begin{equation*}
    \begin{aligned}
 D_{\text{JS}}(\rho_{i}, \rho_{j}) &= \frac{1}{2}\mathbb{E}_{\rho_{i} (s,a)} \log\frac{\zeta_{ij}(s,a)}{1+\zeta_{ij}(s,a)} + \frac{1}{2}\mathbb{E}_{\rho_{j} (s,a)} \log\frac{1}{1+\zeta_{ij}(s,a)} + \log2
    \end{aligned}
\end{equation*}
\begin{equation*}
    \begin{aligned}
    D_{\text{KLS}}(\rho_{i}, \rho_{j}) &= \mathbb{E}_{\rho_{i} (s,a)} \log\zeta_{ij}(s,a) - \mathbb{E}_{\rho_{j} (s,a)} \log\zeta_{ij}(s,a)
    \end{aligned}
\end{equation*}

The SVPG gradient involves the gradient of the $f$-divergence {\em w.r.t.} the policy parameters ($\theta$). For $D_{\text{JS}}$ and $D_{\text{KLS}}$, the gradient can be written using $\zeta$ as follows:
\begin{equation}\label{eq:divergence_grad}
    \nabla_{\theta_j} D_{\text{JS}} = \nabla_{\theta_j}\mathbb{E}_{\rho_{j}} \underbrace{-(1/2)\log[1+\zeta_{ij}(s,a)]}_{r(s,a)}; \quad \:
    \nabla_{\theta_j} D_{\text{KLS}} =  \nabla_{\theta_j}\mathbb{E}_{\rho_{j}} \underbrace{[-\zeta_{ij}(s,a) - \log\zeta_{ij}(s,a)]}_{r(s,a)}
\end{equation}
The proofs for these are included in Appendix~\ref{appn:gradient_proof}. In practice, these gradients could be estimated using the policy-gradient theorem~\citep{sutton2000policy} with the appropriate term as the reward function (marked as $r(s,a)$ above). It is thus evident that a reasonable estimation of the distribution ratio yields a good approximation of the SVPG gradient (Equation~\ref{eq:svpg_general}), which could then be applied to the policy parameters to learn a QD ensemble. We now discuss methods to estimate $\zeta$ efficiently from samples.

\subsection{Estimating the distribution ratio $\zeta$}
\label{subsec:estimatingRatio}
We start with Noise-Constrastive Estimation (NCE)~\citep{gutmann2010noise} which has found wide applicability in representation learning, natural language processing and image synthesis, among others. We then examine three distribution ratio estimators -- {\em DualDICE}~\citep{nachum2019dualdice} and 
{\em GenDICE}~\citep{zhang2020gendice} were recently proposed for behavior-agnostic off-policy evaluation, and {\em ValueDICE}~\citep{kostrikov2019imitation} enables imitating expert trajectories without requiring additional policy rollouts in the environment.

\noindent
\textbf{NCE.} It provides a method to learn an estimator ${\widetilde{\rho}}_{i}(s,a;\omega)$ for the stationary distribution of any policy $\pi_i$ in the ensemble. NCE uses a noise distribution $p_N(s,a)$ and frames the following binary classification objective:
\[
    \max_\omega \mathbb{E}_{\rho_{i}} \log \frac{{\widetilde{\rho}}_{i}(s,a;\omega)}{{\widetilde{\rho}}_{i}(s,a;\omega) + p_N(s,a)} + \mathbb{E}_{p_N} \log \frac{p_N(s,a)}{{\widetilde{\rho}}_{i}(s,a;\omega) + p_N(s,a)}
\]
\citet{gutmann2010noise} show that under mild assumption on the noise distribution, ${\widetilde{\rho}}_{i}(\cdot;\omega)$ converges to the true density $\rho_{i}$ in the limit of infinite amount of samples. They further note that for practical efficiency, it is desirable to select a noise distribution that is easy to sample from, and that is not too far from the true unknown data distribution $\rho_{i}$. Consequently, for learning the estimator for policy $i$, we use a uniform mixture of stationary distributions of the remaining $(n-1)$ policies in the ensemble as the constrastive noise, {\em i.e.}, $p_N(s,a) = (1/(n-1))\sum_{j \neq i} \rho_{j}(s,a)$. The distribution ratio for a pair of policies can then be computed as $\zeta_{ij}(s,a) = {\widetilde{\rho}}_{i}(s,a) / {\widetilde{\rho}}_{j}(s,a)$.

\noindent
\textbf{{\em DualDICE}.}~\citet{nachum2019dualdice} propose a convex optimization problem that gives the distribution ratio as its optimal solution:
\begin{equation}\label{eq:dualD_basic}
    \zeta_{ij} = \argmin_{x:\mathcal{S}\times\mathcal{A}\rightarrow\mathbb{R}} \frac{1}{2} \mathbb{E}_{(s,a)\sim\rho_{j}} [x(s,a)^2] - \mathbb{E}_{(s,a)\sim\rho_{i}} [x(s,a)]
\end{equation}
The expression is then simplified with the following {\em change-of-variables} trick. Define a variable $\nu(s,a)$ and the operator $\mathcal{B}^{\pi_i}\nu(s,a) = \gamma \mathbb{E}_{s'\sim p(\cdot|s,a), a' \sim \pi_i(s')}[\nu(s',a')]$. Using $x(s,a) = \nu(s,a) - \mathcal{B}^{\pi_i}\nu(s,a)$, the second expectation in Equation~\ref{eq:dualD_basic} telescopes and conveniently reduces into an expectation over the initial states. The transformed objective is:
\[
    \min_{\nu:\mathcal{S}\times\mathcal{A}\rightarrow\mathbb{R}} \frac{1}{2} \mathbb{E}_{(s,a)\sim\rho_{j}}[(\nu - \mathcal{B}^{\pi_i}\nu)(s,a)^2] - (1-\gamma) \mathbb{E}_{\substack{s_0 \sim \mu_0 \\ a_0 \sim \pi_i(s_0)}}[\nu(s_0,a_0)]
\]
Given an optimal solution $\nu^*$ for this equation, the distribution ratio is recovered with $\zeta_{ij}(s,a) = (\nu^* - \mathcal{B}^{\pi_i}\nu^*)(s,a)$. Further, to alleviate the bias in the sample-based Monte-Carlo estimate of the gradient,~\citet{nachum2019dualdice} suggest the use of Fenchel conjugates. The final objective is a min-max saddle-point optimization that directly provides the distribution ratio $\zeta_{ij}(s,a)$. Appendix~\ref{appn:dualdice} includes the details.

\noindent
\textbf{{\em ValueDICE}.} The Donsker-Varadhan representation~\citep{donsker1983asymptotic} of the KL-divergence is given by: 
\[
    D_{\text{KL}}(\rho_{i}||\rho_{j}) = \sup_{x:\mathcal{S}\times\mathcal{A}\rightarrow\mathbb{R}} \mathbb{E}_{(s,a)\sim\rho_{i}} [x(s,a)] - \log \mathbb{E}_{(s,a)\sim\rho_{j}} [e^{x(s,a)}]
\]
In {\em ValueDICE}~\citep{kostrikov2019imitation}, the authors use the fact that the optimality in the above equation is achieved at $x^*(s,a) = \log \zeta_{ij}(s,a) + C$, for some constant $C \in \mathbb{R}$. The proof is included in Appendix~\ref{appn:valuedice} for completeness. Therefore, a method to obtain $\zeta_{ij}$ is to first solve for $x^*$ (written as a minimization):
\begin{equation*}
    x^* = \argmin_{x:\mathcal{S}\times\mathcal{A}\rightarrow\mathbb{R}} \log \mathbb{E}_{(s,a)\sim\rho_{j}} [e^{x(s,a)}] - \mathbb{E}_{(s,a)\sim\rho_{i}} [x(s,a)]
\end{equation*}
With the same {\em change-of-variables} trick from {\em DualDICE}, $x(s,a) = \nu(s,a) - \mathcal{B}^{\pi_i}\nu(s,a)$, the second expectation over $\rho_{i}(s,a)$ is transformed into an expectation over the initial states:
\[
\min_{\nu:\mathcal{S}\times\mathcal{A}\rightarrow\mathbb{R}}  \log \mathbb{E}_{(s,a)\sim\rho_{j}}[e^{\nu(s,a) - \mathcal{B}^{\pi_i}\nu(s,a)}] - (1-\gamma) \mathbb{E}_{\substack{s_0 \sim \mu_0 \\ a_0 \sim \pi_i(s_0)}}[\nu(s_0,a_0)]
\]
Different from {\em DualDICE}, {\em ValueDICE} avoids the min-max saddle-point optimization by eschewing the use of Fenchel conjugates to remove the bias in the sample-based gradient. The log distribution ratio is calculated (up to a constant shift) from $\nu^*$ as, $\log\zeta_{ij}(s,a) = \nu^*(s,a) - \mathcal{B}^{\pi_i}\nu^*(s,a)$.

\noindent
\textbf{{\em GenDICE}.} It is known that the stationary distribution for a policy $\pi_i$ satisfies the following Bellman flow constraint:
\[
    \rho_i(s',a') = (1-\gamma)\mu_0(s')\pi_i(a'|s') + \gamma \int \pi_i(a'|s')p(s'|s,a)\rho_i(s,a) ds da; \quad \forall (s',a') \in \mathcal{S}\times\mathcal{A}
\]
This could be re-written using the distribution ratio $\zeta_{ij}$ as:
\begin{equation}\label{eq:bellman_flow}
     \rho_j(s',a') \zeta_{ij}(s',a') = \underbrace{(1-\gamma)\mu_0(s')\pi_i(a'|s') + \gamma \int \pi_i(a'|s')p(s'|s,a)\zeta_{ij}(s,a) \rho_j(s,a) ds da}_{\mathcal{T}_{(\pi_i, \rho_j)} \circ \zeta_{ij}}
\end{equation}
\citet{zhang2020gendice} parameterize $\zeta_\theta$ and suggest to estimate it by minimizing the $f$-divergence between the distributions (with support on $\mathcal{S} \times \mathcal{A}$) on the two sides of Equation~\ref{eq:bellman_flow}, namely $\rho_j . \zeta_\theta$ and $\mathcal{T}_{(\pi_i, \rho_j)} \circ \zeta_\theta$, where the notation $\mathcal{T}_{(\pi_i, \rho_j)}$ denotes the distribution operator on the RHS in Equation~\ref{eq:bellman_flow}. The objective, which further includes a penalty regularizer on $\zeta_\theta$ to prevent degenerate solutions, is:
\[
 \zeta_{ij} = \argmin_\theta D_f\big(\mathcal{T}_{(\pi_i, \rho_j)} \circ \zeta_\theta \:||\:\rho_j . \zeta_\theta\big) + \frac{\lambda}{2}{(\mathbb{E}_{\rho_j}[\zeta_\theta] - 1)}^2
\]
Similar to {\em DualDICE}, Fenchel conjugates are used to obtain unbiased gradient estimates, resulting in a min-max saddle-point optimization. The final objective, with $\chi^2$ as the $f$-divergence is included in Appendix~\ref{appn:gendice}.

\noindent
\textbf{Overall Algorithm.}. We summarize our complete approach for training a QD policy ensemble in Algorithm~\ref{algo:qd}. In each iteration, we sample transitions in the environment using the policies in the ensemble and update the networks that facilitate estimation of the distribution ratios $\zeta_{ij}$. The type of network(s) and the update rule is determined by the estimator choice. To form the SVPG gradient (Equation~\ref{eq:svpg_general}), the current value of $\zeta$ is used to compute the divergence value and the divergence gradient. The latter, as shown by Equation~\ref{eq:divergence_grad}, is equivalent to the policy gradient with a distinctive reward function. We use the clipped PPO algorithm~\citep{ppo} for the policy gradient, although other on-policy and off-policy RL methods are also applicable.  

\RestyleAlgo{ruled}
\LinesNumbered
\begin{algorithm}[t]
\small
\SetNoFillComment
\newcommand\mycommfont[1]{\footnotesize\sffamily\textcolor{blue}{#1}}
\SetCommentSty{mycommfont}

\DontPrintSemicolon
\SetKwComment{Comment}{$\triangleright$\ }{}
 \BlankLine
Initialize policy ensemble $\{\pi_i\}^n_{1}$ \\
Initialize networks to estimate $\zeta_{ij}$ \Comment*[r]{Parameterization depends on the method (Section~\ref{subsec:estimatingRatio})} 

 \BlankLine

 \For{each iteration}{
     Sample rollouts with $\pi_i \:\: \forall i$ \\
     
     Update all $\zeta_{ij}$ estimation networks \Comment*[r]{Objective depends on the method (Section~\ref{subsec:estimatingRatio})} 
        
     Use $\zeta_{ij}$ to compute the divergence value and divergence gradient \Comment*[r]{($D_{\text{JS}}$ or $D_{\text{KLS}}$)} 
     Update each $\pi_i$ with the corresponding SVPG gradient \Comment*[r]{(Equation~\ref{eq:svpg_general})}  
 }
 \caption{Pseudo code for learning a QD ensemble}
 \label{algo:qd}
\end{algorithm}

\section{Related work}
Neuroevolution methods inspired by Quality-Diversity~\citep{pugh2016quality} have been proposed to efficiently manage the exploration-exploitation trade-off in RL.~\citet{conti2017improving} augment evolution strategies~\citep{salimans2017evolution} such that the fitness of a particle is computed by a weighted combination of the performance and novelty components. The novelty is determined based on a chosen behavior characterization (BC) metric. In MAP-Elites~\citep{mouret2015illuminating}, the entire behavior space is divided into a discrete grid, where each grid-dimension represents a BC. The algorithm then fills each grid cell with the highest quality solution possible for that cell. Recent methods integrate RL gradients with concepts from evolutionary computation ({\em e.g. random mutations}) to learn diverse exploratory agents~\citep{khadka2019collaborative, liu2019emergent} or to discover coordination strategies for multi-agent RL~\citep{khadka2019evolutionary}.

\textbf{Diversity Maximization in RL.} To aid exploration in sparse-reward tasks, ~\citet{hong2018diversity} encourage the current policy to be diverse compared to an archive of past policies, by maximizing a distance metric in the action space. Expanding on this idea,~\citet{doan2019attraction} ensure sufficient diversity in a population by using {\em operators} for attraction and repulsion between agents. Towards learning diverse skills even in the absence of an external reward signal, maximization of the mutual information between the latent skill and the state-visitation of the skill-conditioned policy has been proposed~\citep{florensa2017stochastic, eysenbach2018diversity}. This is achieved by training a neural network to estimate the latent skill posterior, which provides proxy rewards for the policy. Our work aims to broaden the SVPG algorithm~\citep{liu2017stein} for learning a QD policy ensemble. We substitute the parameter-space RBF kernel used in their method with negative exponents of $f$-divergences, and employ distribution ratio estimation techniques to approximate the ensuing gradient on the policy parameters.~\citet{gangwani2018learning} avail SVPG to improve self-imitation learning. Their procedure bears some resemblance to our NCE-based ratio estimation, though, in contrast to them, we use a mixture of stationary distributions as the contrastive noise.

        
    
        

\section{Experiments}
\label{sec:exp}
In this section, we train policy ensembles in various environments with continuous state- and action-space. We evaluate the different distribution ratio estimators and divergence metrics from Section~\ref{sec:methods}. For ease of exposition, the algorithms are abbreviated as \textbf{QD-{\em \{ratio-estimator\}}-{\em \{divergence\}}}, hence QD-NCE-JS, for instance, is Algorithm~\ref{algo:qd} instantiated with $\exp (-D_{\text{JS}}(\rho_j, \rho_i)/\mathit{T})$ as the kernel for SVPG, and NCE as the estimator for $\zeta$. Our goal is to 
gauge the effectiveness of our approach in producing diverse, high-quality behaviors and suitably handling tasks with deceptive rewards. We also compare the NCE and DICE-based estimators on a quantitative metric correlated with behavioral diversity. The hyperparameters and implementation details are included in Appendix~\ref{appn:hyperparameters}.

\begin{figure}[t]
\centering
\captionsetup[subfigure]{justification=centering}
    \begin{subfigure}[t]{0.15\textwidth}
        \centering
        \includegraphics[scale=0.08]{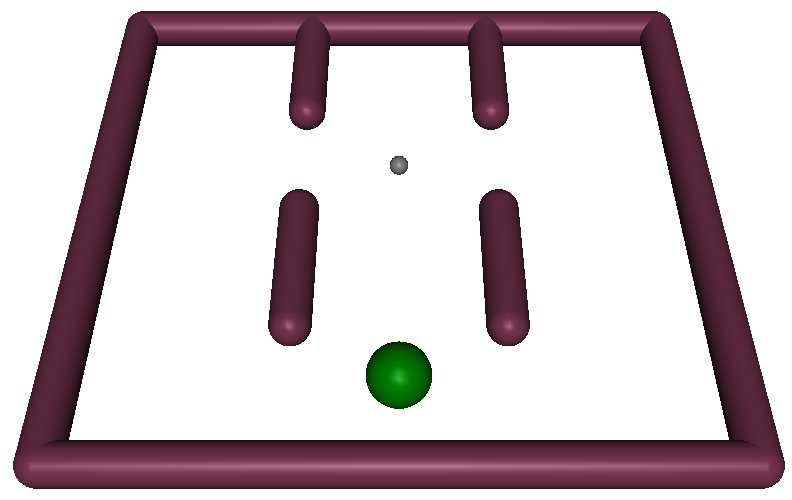}
        \caption{}
        \label{fig:mazeEnv}
    \end{subfigure}\hfill
    \begin{subfigure}[t]{0.15\textwidth}
        \centering
        \includegraphics[scale=0.15]{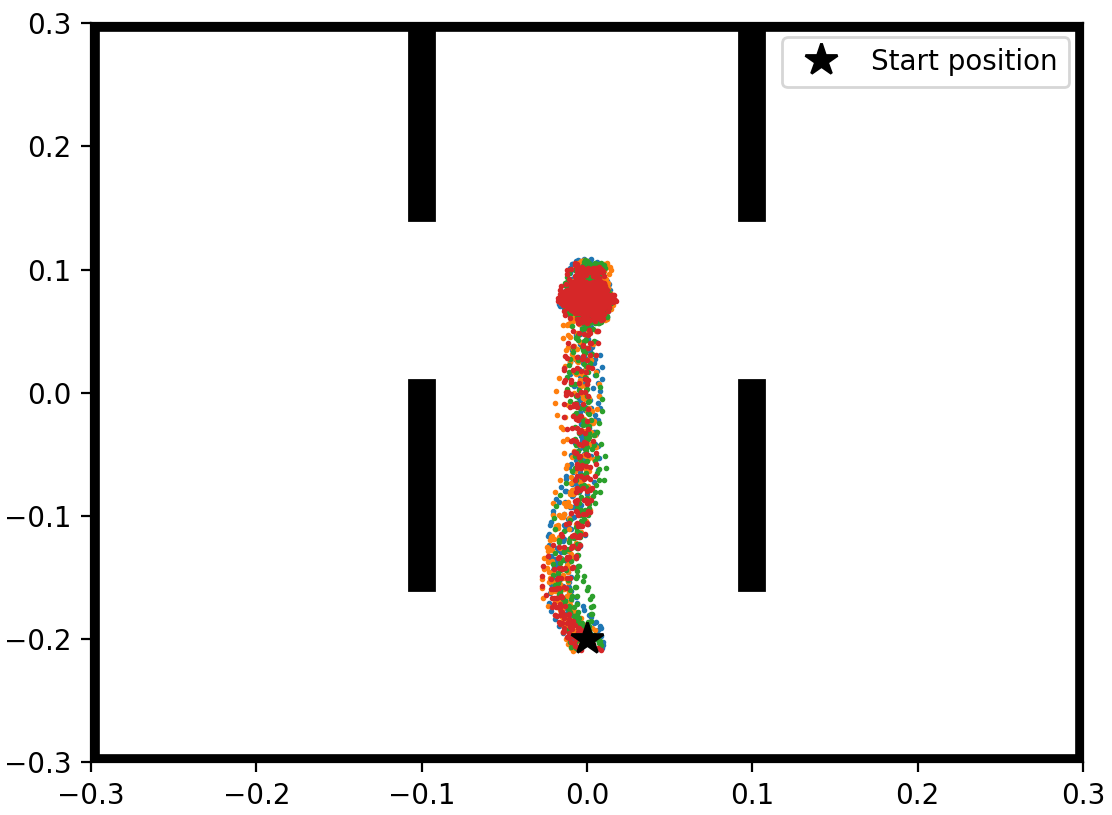}
        \caption{Standard RL}
        \label{fig:maze_base}
    \end{subfigure}\hfill
    \begin{subfigure}[t]{0.15\textwidth}
        \centering
        \includegraphics[scale=0.15]{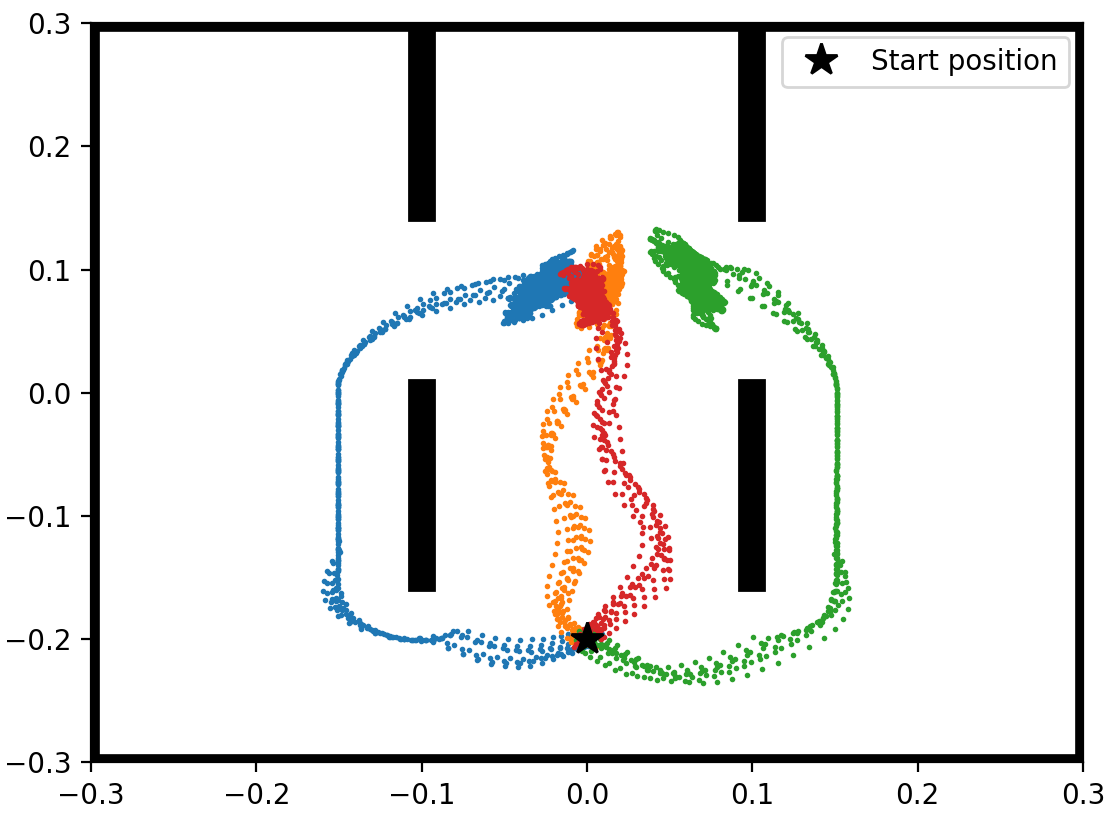}
        \caption{QD-NCE-JS}
        \label{fig:maze_nice_js}
    \end{subfigure}\hfill
    \begin{subfigure}[t]{0.15\textwidth}
        \centering
        \includegraphics[scale=0.15]{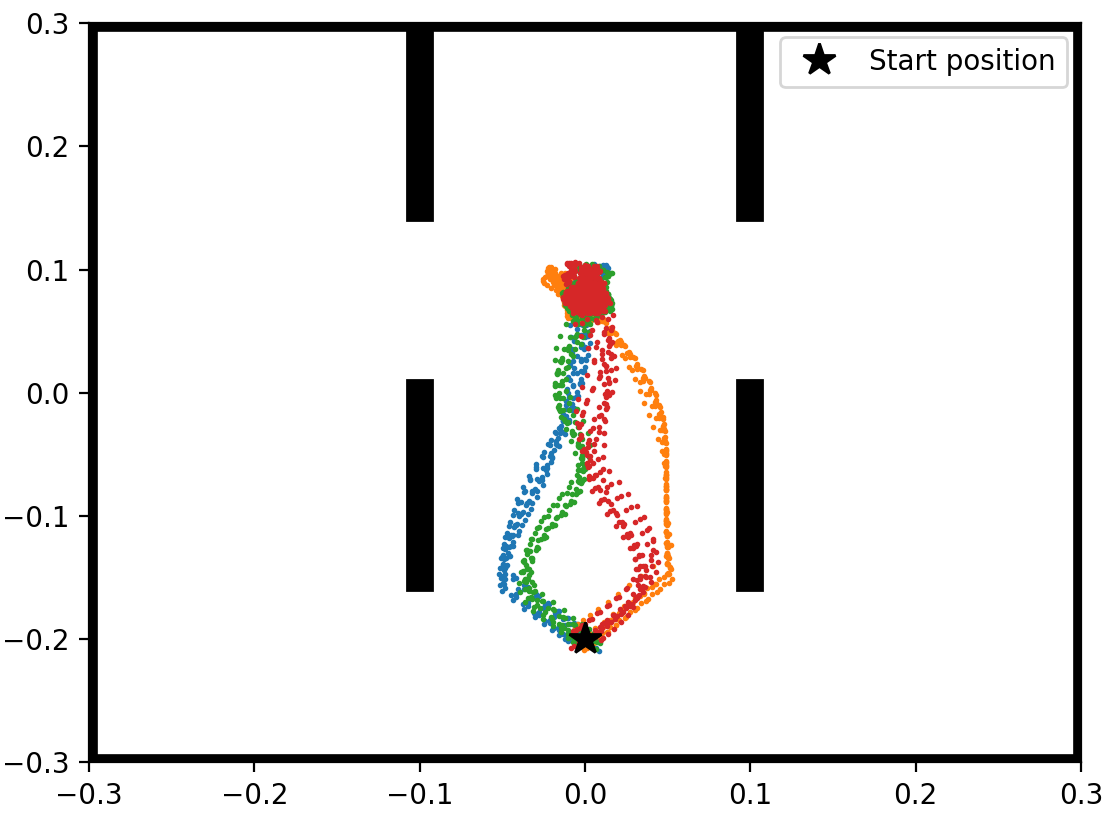}
        \caption{QD-NCE-KLS}
    \end{subfigure}\hfill
    \begin{subfigure}[t]{0.15\textwidth}
        \centering
        \includegraphics[scale=0.15]{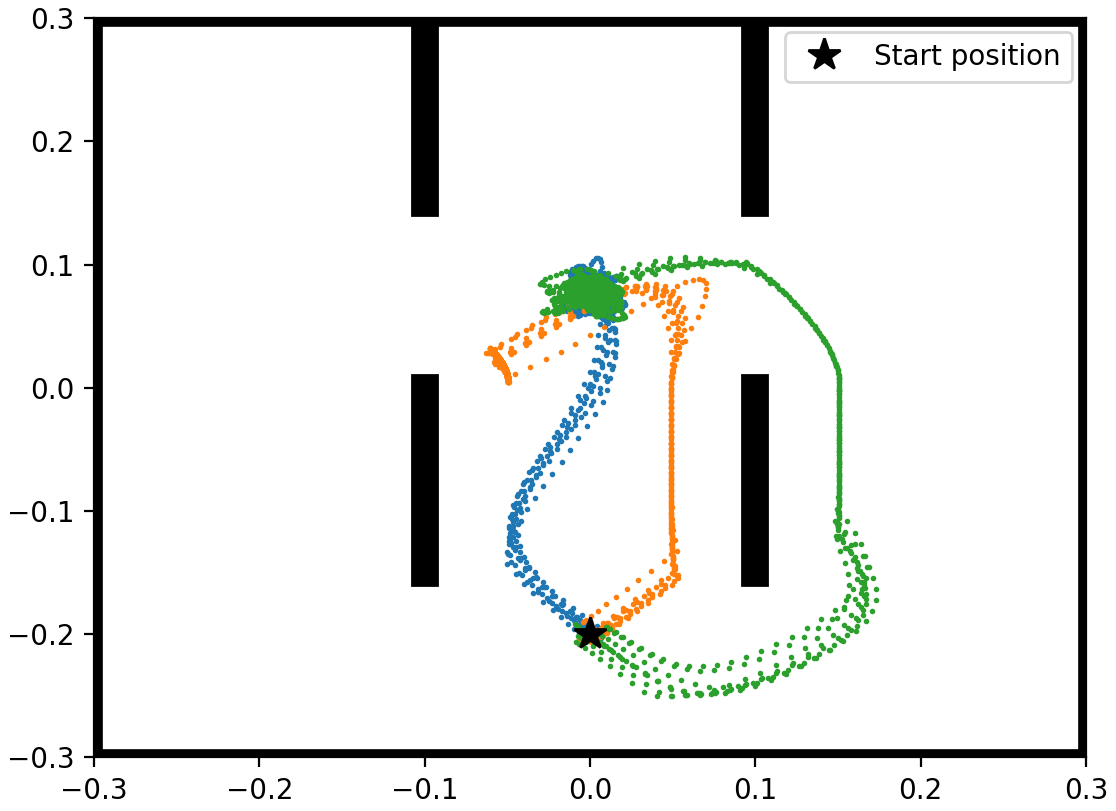}
        \caption{QD-{\em GenDICE}-JS}
    \end{subfigure}\hfill
    \begin{subfigure}[t]{0.15\textwidth}
        \centering
        \includegraphics[scale=0.15]{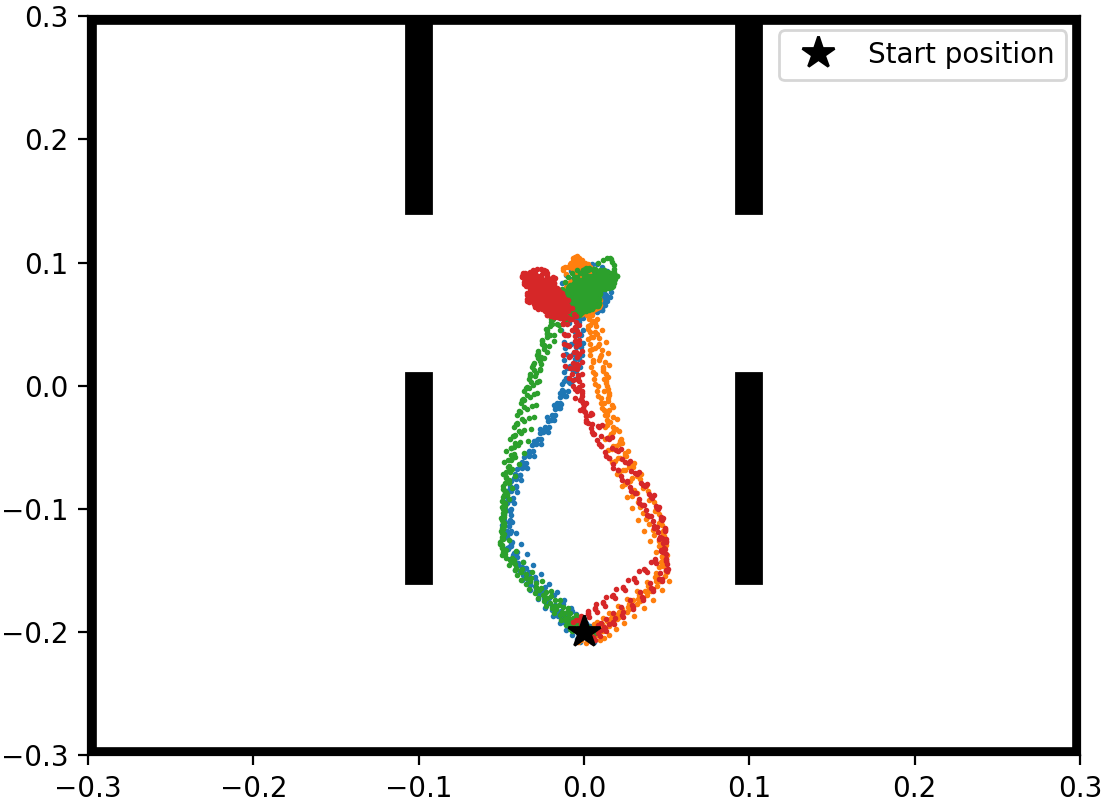}
        \caption{QD-{\em GenDICE}-KLS}
        \label{fig:maze_gd_kls}
    \end{subfigure}\hfill
    \caption{2D Maze navigation task along with trajectories (state-visitations) for several methods.}
    \label{fig:maze_overall}
\end{figure}

\textbf{Qualitative Assessment of QD Behavior.} We visualize the diversity of the learned skills in two environments -- a robotic manipulator arm~\citep{chen2018hardware} and a 2D maze goal navigation task. The robotic arm models a 7 DOF Sawyer robot and is implemented in MuJoCo. For the peg-insertion task, we train a QD ensemble of 10 policies using the exponential of the negative Euclidean distance between the peg and the hole as the per-step reward for RL (the quality measure). Figure~\ref{fig:sawyer_dd_js} shows some of the policies from a single ensemble trained with QD-{\em DualDICE}-JS. We find that while all the policies insert the peg in the hole, the final positions of the joints (marked with white rings) and the end-effector are markedly different. The resultant behaviors have dissimilar torque demands on the various joints, which is advantageous in scenarios such as transfer learning to a robot with system dynamics discrepancies. Figure~\ref{fig:mazeEnv} depicts a 2D navigation task with the start position (green ball at center bottom) and the goal location (small grey circle in the center of the maze). The per-step RL reward is the exponential of the negative Euclidean distance to the goal. We train an ensemble of 10 policies and plot final trajectories from some of them (each policy colored differently). Figure~\ref{fig:maze_base} shows results with the standard RL method, {\em i.e}, no diversity enforcement; the trajectories achieve the best possible cumulative returns but exhibit identical behavior. Figure~\ref{fig:maze_nice_js}-~\ref{fig:maze_gd_kls} plots the paths for policies learned with the QD algorithm (specific instantiation mentioned in the caption). Though the cumulative returns now are lower than those with standard RL, the policies are noticeably more exploratory and cover large portions of the state-space.

\begin{figure}[t]
\centering
\captionsetup[subfigure]{justification=centering}
    \begin{subfigure}[t]{0.2\textwidth}
        \centering
        \hspace*{-0.8cm}
        \includegraphics[scale=0.25]{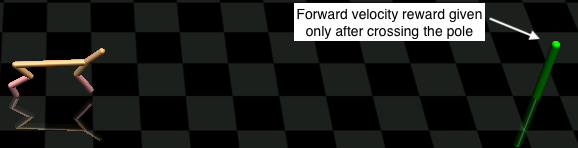}
        \caption{}
        \label{fig:cheetah_env}
    \end{subfigure}\hfill
    \begin{subfigure}[t]{0.78\textwidth}
        \centering
        \hspace*{1cm}
        \includegraphics[scale=0.3]{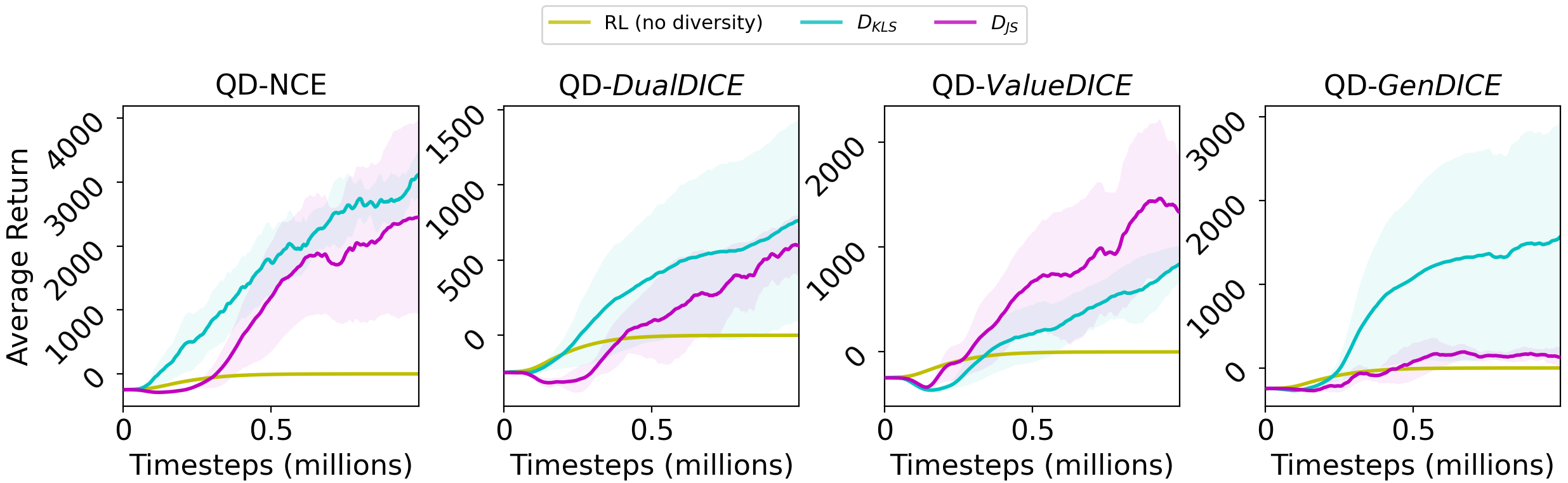}
        \caption{}
         \label{fig:cheetah_perf}
    \end{subfigure}\hfill
    \caption{(a) Modified Half-Cheetah task that introduces multi-modality due to deceptive rewards; (b) Contrasting performance of standard RL (no diversity) with QD method in Algorithm~\ref{algo:qd}.}
    \label{fig:cheetah_overall}
\end{figure}

\textbf{Multi-modal Locomotion with Deceptive Rewards.} One of the crucial benefits of learning a QD ensemble is that it potentially avoids the local optimum trap in the policy-search landscape due to deceptive rewards -- if one policy gets stuck, the explicit diversity enforcement prevents other policies in the ensemble from the same fate. We evaluate this hypothesis with the Half-Cheetah locomotion task from OpenAI Gym~\citep{brockman2016openai}. We modify the task such that the forward velocity reward is only given to the agent once the center-of-mass of the bot is beyond a certain threshold distance ($d$). Concretely, $r_t = vel_{x(t)} *\mathbbm{1}(pos_{x(t)} >= d) - 0.1*\| a_t\|^2_2$, where the second term penalizes large actions and is the default from Gym. Figure~\ref{fig:cheetah_env} is a rendering of the task. This change introduces multi-modality for policy optimization with a locally optimal solution to stand still at the starting location to avoid any action penalty. We compare the performance of the QD ensembles with a baseline standard-RL ensemble. The standard-RL ensemble has the same size as others but the constituent policies do not have any interactions; they apply independently computed gradients. For all baseline and QD ensembles, we select the policy with the highest cumulative returns after training and plot its learning curve in Figure~\ref{fig:cheetah_perf}. We observe that the baseline RL (no diversity) latches onto the deceptive reward of minimizing the action penalty and gets stuck, achieving a cumulative return close to zero. In contrast, the diversity enforcing mechanism in the QD* ensemble enables {\em at-least} one member to reach the alternative mode where high forward velocity rewards are attained. This is evident in the final score accumulated by the member selected from each ensemble.

\begin{wraptable}[12]{r}{3.75cm}
\vspace{-7mm}
\resizebox{0.29\columnwidth}{!}{
\begin{tabular}{ccc}
 & \multicolumn{2}{c}{Hist. Variance $\uparrow$} \\ \cline{2-3} 
Method & Walker-2d & Hopper \\ \hline
\multicolumn{1}{c|}{QD-DD-JS} & \textbf{1.36} & 0.45 \\
\multicolumn{1}{c|}{QD-VD-JS} & 1.33 & \textbf{0.50} \\
\multicolumn{1}{c|}{QD-GD-JS} & 0.63 & 0.14 \\
\multicolumn{1}{c|}{QD-NCE-JS} & 0.13 & 0.11 \\
\multicolumn{1}{c|}{QD-DD-KLS} & 0.10 & 0.10 \\
\multicolumn{1}{c|}{QD-VD-KLS} & 0.24 & 0.45 \\
\multicolumn{1}{c|}{QD-GD-KLS} & 0.07 & 0.40 \\
\multicolumn{1}{c|}{QD-NCE-KLS} & 0.14 & 0.28 \\
\multicolumn{1}{c|}{\citet{gangwani2018learning}} & 0.10 & 0.08 \\
\multicolumn{1}{c|}{DIAYN~\citep{eysenbach2018diversity}} & 0.22 & 0.11
\end{tabular}}
\vspace{-1.5mm}
\caption{\small{Diversity metric (histogram variance) with different estimators. Higher is better. \scriptsize{Mnemonic: DD={\em DualDICE}, VD={\em ValueDICE}, GD=\em{GenDICE}}}}\label{tab:hist_var}
\end{wraptable}
\textbf{Quantitative Comparison of the Estimators.} While the previous experiment exhibits that the NCE and DICE-based estimators can provide adequate diversity impetus, it does not provide insights about the comparative efficiency of the estimators in generating behavioral diversity in the trained ensemble. This is because the forward velocity reward is a {\em quality metric}, which is usually not aligned with the measure of diversity. For instance, an estimator may produce a policy that makes the Half-Cheetah run backwards---this is much desired from the diversity perspective but would perform badly on the quality metric that rewards forward motion. To evaluate the efficacy of our estimators for producing diverse behaviors, and also for meaningful comparison with prior work~\citep{eysenbach2018diversity,gangwani2018learning}, we define a {\em diversity metric} as follows. For two locomotion tasks from Gym (Hopper and Walker-2d), we train policy ensembles without any environmental rewards. Thus, the gradient from the quality-enforcing component in Equation~\ref{eq:svpg} is absent and the QD ensemble is trained only to maximize diversity. Post-training, we generate a few trajectories with all the constituent policies and plot a histogram with the velocity of the center-of-mass of the bot on the $x$-axis and the respective counts on the $y$-axis. We define the diversity metric to be the {\em variance} of this histogram. Intuitively, higher variance in the velocity of the bot is indicative of stronger behavioral diversity in the trained ensemble. Table~\ref{tab:hist_var} evaluates the various estimator on this diversity metric. We note that DICE-based estimators generally outperform NCE. Our intuition for this observation is that since NCE is an on-policy estimator (in contrast with the DICE-based estimators, which are off-policy), the availability of limited on-policy data in each iteration of Algorithm~\ref{algo:qd} has an impact on the efficiency of NCE. Lastly, many of the QD* methods compare favorably to the prior methods for learning diverse skills without environmental rewards~\citep{eysenbach2018diversity,gangwani2018learning}.

\section{Conclusion and Future Work}
In this paper, we study methods to learn diverse and high-return policies. We extend the kernel-based SVPG algorithm with kernels based on $f$-divergence between the stationary distributions of policies. For kernels based on $D_\text{JS}$ and $D_\text{KLS}$, we show that the problem reduces to that of efficient estimation of the ratio of the stationary distributions between policies. To compute these ratios, and consequently the SVPG gradient, we harness noise-contrastive estimation and several distribution ratio estimators widely used for off-policy evaluation and imitation learning. Experimental evaluation with continuous state- and action-space environments demonstrates that the approach is capable of generating diverse high-quality skills, assists in multi-modal environments with deceptive rewards, and provides a constructive learning signal when the external rewards are absent. Our algorithmic framework is general enough to accommodate any distribution ratio estimator. Utilizing future research on these estimators for improving the efficiency of QD training is an interesting direction, along with investigating which other $f$-divergences or integral probability metrics (IPMs such as the Wasserstein distance and the Maximum Mean Discrepancy) between stationary distributions could be incorporated into the framework.



\clearpage
\acknowledgments{This work is supported by the National Science Foundation under grants OAC-1835669 and CCF-2006526. Yuan Zhou is supported in part by a Ye Grant and a JPMorgan Chase AI Research Faculty Research Award.}


\bibliography{main}  

\clearpage
\appendix

\section{Appendix}
\subsection{Gradient of divergences {\em w.r.t.} the policy parameters}\label{appn:gradient_proof}
We derive the expressions for $\nabla_{\theta_j} D_{\text{JS}}$ and $\nabla_{\theta_j} D_{\text{KLS}}$ mentioned in Equation~\ref{eq:divergence_grad}. $\theta_j$ denotes the parameters for $\pi_j$. The distribution ratio, $\zeta_{ij} = \rho_{i}/\rho_{j}$, depends on $\theta_j$ through $\rho_{j}$. A bar above a symbol signifies that it is a constant {\em w.r.t.} $\theta_j$; for instance, while $\rho_{j}$ depends on $\theta_j$, $\bar{\rho}_{j}$ does not. The derivation uses the property that the expectation of the score function estimator is 0.

\textbf{Jenson-Shannon divergence.}
\begin{equation*}
 D_{\text{JS}}(\rho_{i}, \rho_{j}) = \frac{1}{2}\mathbb{E}_{\rho_{i}} \log\frac{\rho_{i}}{\rho_{i}+\rho_{j}} + \frac{1}{2}\mathbb{E}_{\rho_{j}} \log\frac{\rho_{j}}{\rho_{i}+\rho_{j}} + \log2 \\
\end{equation*}
Differentiating with the product rule,
\begin{equation*}
\begin{aligned}
    \nabla_{\theta_j} 2D_{\text{JS}} =& - \mathbb{E}_{\rho_{i}} \nabla_{\theta_j} \log [\rho_{i}+\rho_{j}] + \underbrace{\mathbb{E}_{\rho_{j}} \nabla_{\theta_j} \log [\rho_{j}]}_{
    \substack{=0 \\ \text{\small{Exp. score function}}}} + \nabla_{\theta_j} \mathbb{E}_{\rho_{j}}  \log [\bar{\rho}_{j}] - \mathbb{E}_{\rho_{j}} \nabla_{\theta_j} \log [\rho_{i}+\rho_{j}] - \nabla_{\theta_j} \mathbb{E}_{\rho_{j}}  \log [\rho_{i}+\bar{\rho}_{j}] \\
    =& - \underbrace{\mathbb{E}_{\rho_{i}+\rho_{j}} \nabla_{\theta_j} \log [\rho_{i}+\rho_{j}]}_{
    \substack{=0 \\ \text{\small{Exp. score function}}}} + \nabla_{\theta_j} \mathbb{E}_{\rho_{j}}  \log [\bar{\rho}_{j}] - \nabla_{\theta_j} \mathbb{E}_{\rho_{j}}  \log [\rho_{i}+\bar{\rho}_{j}] \\
    \nabla_{\theta_j} D_{\text{JS}} =& -(1/2) \nabla_{\theta_j}\mathbb{E}_{\rho_{j}} \log[1+\bar{\zeta}_{ij}]
  \end{aligned}  
\end{equation*}

\textbf{Symmetric Kullback-Leibler divergence.}
\begin{equation*}
    D_{\text{KLS}}(\rho_{i}, \rho_{j}) = \mathbb{E}_{\rho_{i}} \log\frac{\rho_{i}}{\rho_{j}} - \mathbb{E}_{\rho_{j}} \log\frac{\rho_{i}}{\rho_{j}}
\end{equation*}

Differentiating with the product rule,
\begin{equation*}
    \nabla_{\theta_j} D_{\text{KLS}} = -\mathbb{E}_{\rho_{i}} \nabla_{\theta_j} \log [\rho_{j}]
    -\nabla_{\theta_j} \mathbb{E}_{\rho_{j}}  \log \bar{\zeta}_{ij} + \underbrace{\mathbb{E}_{\rho_{j}}  \nabla_{\theta_j} \log[\rho_{j}]}_{
    \substack{=0 \\ \text{\small{Exp. score function}}}} 
\end{equation*}
For the first term, interchanging the gradient and the expectation, we can write:
\begin{equation*}
    \mathbb{E}_{\rho_{i}} \nabla_{\theta_j} \log [\rho_{j}] = \sum_{(s,a)} \rho_{i} \frac{\nabla_{\theta_j} \rho_{j}}{\bar{\rho}_{j}} = \sum_{(s,a)} \bar{\zeta}_{ij} \nabla_{\theta_j} \rho_{j} = \nabla_{\theta_j} \mathbb{E}_{\rho_{j}} [\bar{\zeta}_{ij}]
\end{equation*}
Therefore,
\begin{equation*}
    \nabla_{\theta_j} D_{\text{KLS}} = \nabla_{\theta_j}\mathbb{E}_{\rho_{j}} [-\bar{\zeta}_{ij} - \log\bar{\zeta}_{ij}]
\end{equation*}

\subsection{{\em DualDICE} min-max objective with Fenchel conjugates}\label{appn:dualdice}
We start with the {\em DualDICE} objective from Section~\ref{subsec:estimatingRatio}:
\[
    J(\nu) = \frac{1}{2} \mathbb{E}_{(s,a)\sim\rho_{j}}[(\nu - \mathcal{B}^{\pi_i}\nu)(s,a)^2] - (1-\gamma) \mathbb{E}_{\substack{s_0 \sim \mu_0 \\ a_0 \sim \pi_i(s_0)}}[\nu(s_0,a_0)]
\]
Fenchel duality provides that $\frac{1}{2}x^2=\max_g gx - \frac{1}{2}g^2$ for a scalar $g \in \mathbb{R}$.~\citet{nachum2019dualdice} rewrite the quadratic (first) term in the objective using this maximization and use the interchangeability principle~\citep{shapiro2014lectures} to replace the inner max over scalar $g$ to a max over functions $g:\mathcal{S}\times\mathcal{A}\rightarrow\mathbb{R}$. Given the definition of the $\mathcal{B}^{\pi_i}$ operator, this yields the min-max objective:
\begin{equation*}
\begin{aligned}
    \min_\nu \max_g J(\nu, g) = \mathbb{E}_{\substack{(s,a)\sim\rho_{j}, s'\sim p(\cdot|s,a) \\ a' \sim \pi_i(s')}} \bigg[\big(\nu(s,a) &- \gamma \nu(s',a')\big)g(s,a) - \frac{g(s,a)^2}{2}\bigg]  \\ 
    &- (1-\gamma) \mathbb{E}_{\substack{s_0 \sim \mu_0 \\ a_0 \sim \pi_i(s_0)}}[\nu(s_0,a_0)] 
\end{aligned}
\end{equation*}
The distribution ratio is obtained from the saddle-point solution $(\nu^*,g^*)$ using the following equivalence, $\zeta_{ij}(s,a) = g^*(s,a) = (\nu^* - \mathcal{B}^{\pi_i}\nu^*)(s,a)$.

\subsection{Optimality in the Donsker-Varadhan representation}\label{appn:valuedice}
The Donsker-Varadhan representation~\citep{donsker1983asymptotic} of the KL-divergence is given by: 
\[
    D_{\text{KL}}(\rho_{i}||\rho_{j}) = \sup_{x:\mathcal{S}\times\mathcal{A}\rightarrow\mathbb{R}} \mathbb{E}_{(s,a)\sim\rho_{i}} [x(s,a)] - \log \mathbb{E}_{(s,a)\sim\rho_{j}} [e^{x(s,a)}]
\]
The optimality is achieved at $x^*(s,a) = \log \zeta_{ij}(s,a) + C$, for some constant $C \in \mathbb{R}$.

\begin{proof}
We begin with a re-write of the expression inside the supremum:
\begin{equation*}
    \begin{aligned}
    &\mathbb{E}_{(s,a)\sim\rho_{i}} \bigg(\log \big[e^{x(s,a)}.\frac{\rho_{i}}{\rho_{j}}.\frac{\rho_{j}}{\rho_{i}}\big]\bigg) - \log \mathbb{E}_{(s,a)\sim\rho_{j}} [e^{x(s,a)}] \\ 
    =& \underbrace{\mathbb{E}_{(s,a)\sim\rho_{i}} [\log \frac{\rho_{i}}{\rho_{j}}]}_{\text{KL}} + \mathbb{E}_{(s,a)\sim\rho_{i}} \bigg( \log\big[\frac{\rho_{j}}{\rho_{i}}.e^{x(s,a)}\big]\bigg) - \log \mathbb{E}_{(s,a)\sim\rho_{j}} [e^{x(s,a)}] \\
    \le& \: D_{\text{KL}}(\rho_{i}||\rho_{j}) +  
    \log\mathbb{E}_{(s,a)\sim\rho_{i}}\big[\frac{\rho_{j}}{\rho_{i}}.e^{x(s,a)}\big] - \log \mathbb{E}_{(s,a)\sim\rho_{j}} [e^{x(s,a)}] \quad\quad \text{\small{(Jensen's inequality)}}\\
    =& \: D_{\text{KL}}(\rho_{i}||\rho_{j}) +  
    \log\mathbb{E}_{(s,a)\sim\rho_{j}}[e^{x(s,a)}] - \log \mathbb{E}_{(s,a)\sim\rho_{j}} [e^{x(s,a)}] \\
    =& \: D_{\text{KL}}(\rho_{i}||\rho_{j})
\end{aligned}
\end{equation*}
Therefore, this expression is upper bounded by $D_{\text{KL}}(\rho_{i}||\rho_{j})$. To complete the proof, we show that this upper bound is indeed achieved when $x(s,a) = \log \zeta_{ij}(s,a) + C$, for some constant $C \in \mathbb{R}$. Inserting this into the expression, we get:
\begin{equation*}
\begin{aligned}
    &\mathbb{E}_{(s,a)\sim\rho_{i}} [\log \zeta_{ij}(s,a) + C] - \log \mathbb{E}_{(s,a)\sim\rho_{j}} [e^{\log \zeta_{ij}(s,a) + C}] \\
    =& \: D_{\text{KL}}(\rho_{i}||\rho_{j}) + C - \log \big(e^C \underbrace{\mathbb{E}_{(s,a)\sim\rho_{j}} [\zeta_{ij}(s,a)]}_{\text{=1}} \big) \\
    =& \: D_{\text{KL}}(\rho_{i}||\rho_{j})
\end{aligned}
\end{equation*}
\end{proof}

\subsection{{\em GenDICE} min-max objective with Fenchel conjugates}\label{appn:gendice}
We start with the {\em GenDICE} objective from Section~\ref{subsec:estimatingRatio} that minimizes the $f$-divergence between the quantities on the two sides of the Bellman flow constraint, along with a penalty regularization:
\[
 J(\theta) = D_f\big(\mathcal{T}_{(\pi_i, \rho_j)} \circ \zeta_\theta \:||\: \rho_j . \zeta_\theta \big) + \frac{\lambda}{2}{(\mathbb{E}_{\rho_j}[\zeta_\theta] - 1)}^2
\]
where $\mathcal{T}_{(\pi_i, \rho_j)}$ is an operator such that: 
\[
(\mathcal{T}_{(\pi_i, \rho_j)} \circ \zeta_\theta) (s',a') \coloneqq (1-\gamma)\mu_0(s')\pi_i(a'|s') + \gamma \int \pi_i(a'|s')p(s'|s,a)\zeta_\theta(s,a) \rho_j(s,a) ds da
\]
Let $g:\mathcal{S}\times\mathcal{A}\rightarrow\mathbb{R}$ be a function. The $f$-divergence could be substituted with its variational representation~\citep{nguyen2010estimating} which involves the Fenchel conjugate ($f^*$) of the $f$ function in $D_f$:
\[
 D_f\big(\mathcal{T}_{(\pi_i, \rho_j)} \circ \zeta_\theta \:||\: \rho_j . \zeta_\theta \big) = \max_g \: \mathbb{E}_{\mathcal{T}_{(\pi_i, \rho_j)} \circ \zeta_\theta} [g(s,a)] - \mathbb{E}_{\rho_j . \zeta_\theta} [f^*(g(s,a))]
\]
This expression can be simplified by using the definition of the $\mathcal{T}_{(\pi_i, \rho_j)}$ operator. Furthermore, since Fenchel duality provides that $\frac{1}{2}x^2=\max_u ux - \frac{1}{2}u^2$, the quadratic penalty regularization can also be written in form of a max over a scalar variable $u \in \mathbb{R}$. This yields the min-max objective:

\begin{equation*}
\begin{aligned}
   \min_\theta \max_{g,u} J(\theta, g, u) = \: &(1-\gamma) \mathbb{E}_{\mu_0(s)\pi_i(a|s)} [g(s,a)] + \gamma \mathbb{E}_{\substack{(s,a)\sim\rho_{j}, s'\sim p(\cdot|s,a) \\ a' \sim \pi_i(s')}} [\zeta_\theta(s,a) g(s',a')] \\ &- \mathbb{E}_{(s,a)\sim\rho_j} [\zeta_\theta(s,a) f^*(g(s,a))] + \lambda \big(\mathbb{E}_{\rho_j}[u\zeta_\theta(s,a) - u] - \frac{u^2}{2}\big)  
\end{aligned}
\end{equation*}

For a practical instantiation,~\citet{zhang2020gendice} suggest the $\chi^2$ divergence, which is an $f$-divergence with $f(x)=(x-1)^2$ and $f^*(x) = x + \frac{x^2}{4}$

\subsection{Further Experiments}\label{appn:further_exp}
\begin{figure}[t]
\centering
\captionsetup[subfigure]{justification=centering}
    \begin{subfigure}[t]{0.12\textwidth}
        \centering
        \includegraphics[scale=0.1]{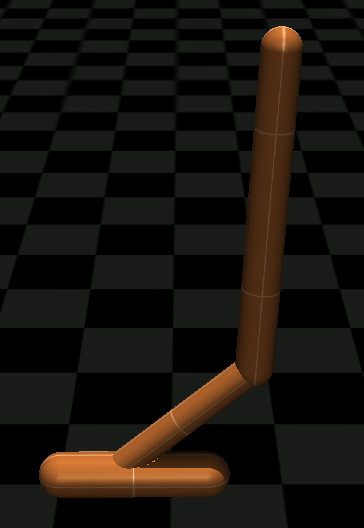}
        \caption{}
        \label{fig:hopper_env}
    \end{subfigure}\hfill
    \begin{subfigure}[t]{0.12\textwidth}
        \centering
        \includegraphics[scale=0.12]{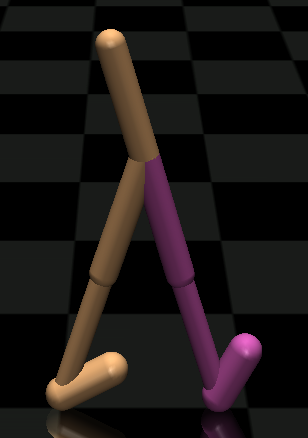}
        \caption{}
        \label{fig:walker_env}
    \end{subfigure}\hfill
    \begin{subfigure}[t]{0.3\textwidth}
        \centering
        \includegraphics[scale=0.22]{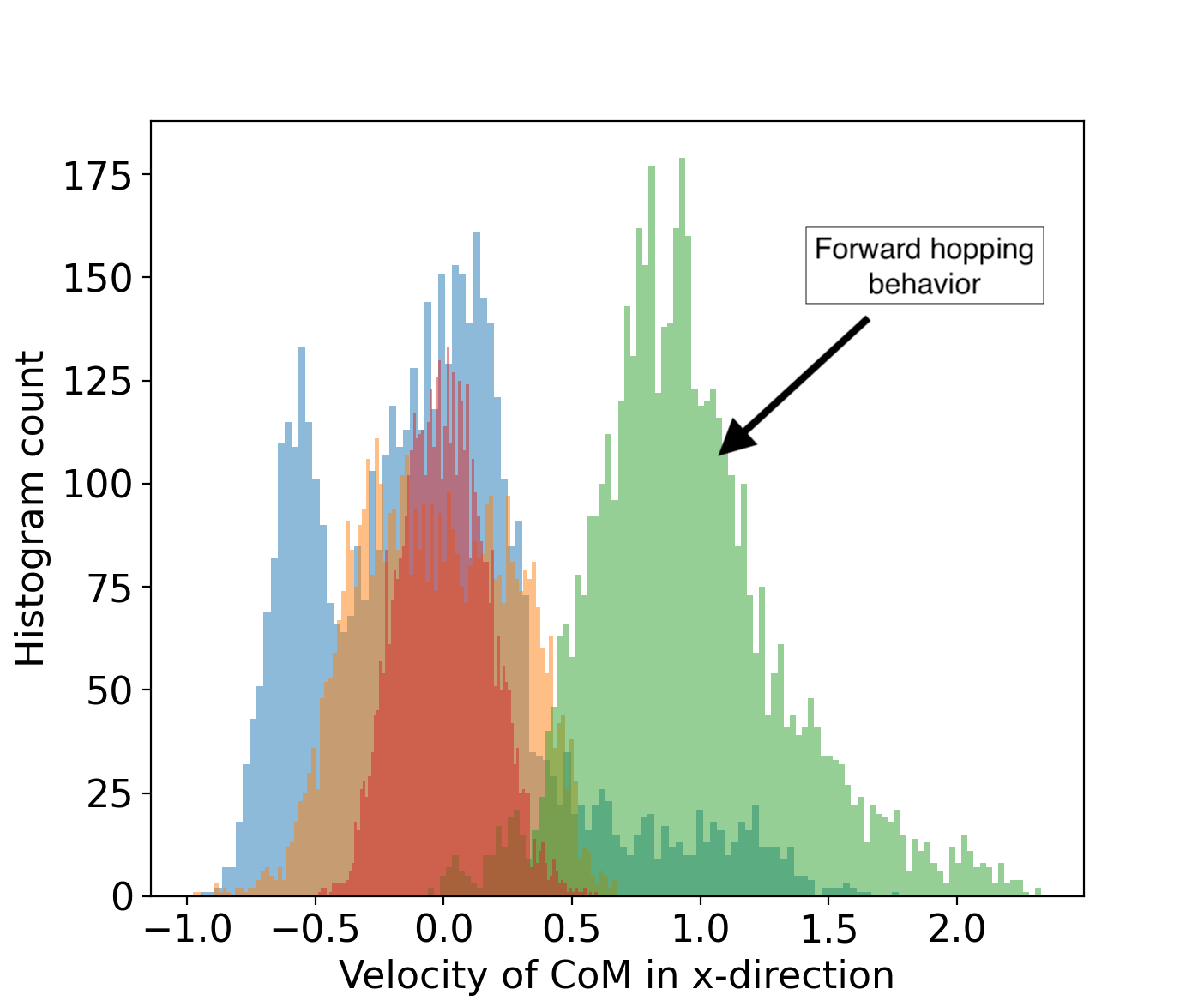}
        \caption{}
        \label{fig:hopper_hist}
    \end{subfigure}\hfill
    \begin{subfigure}[t]{0.3\textwidth}
        \centering
        \includegraphics[scale=0.22]{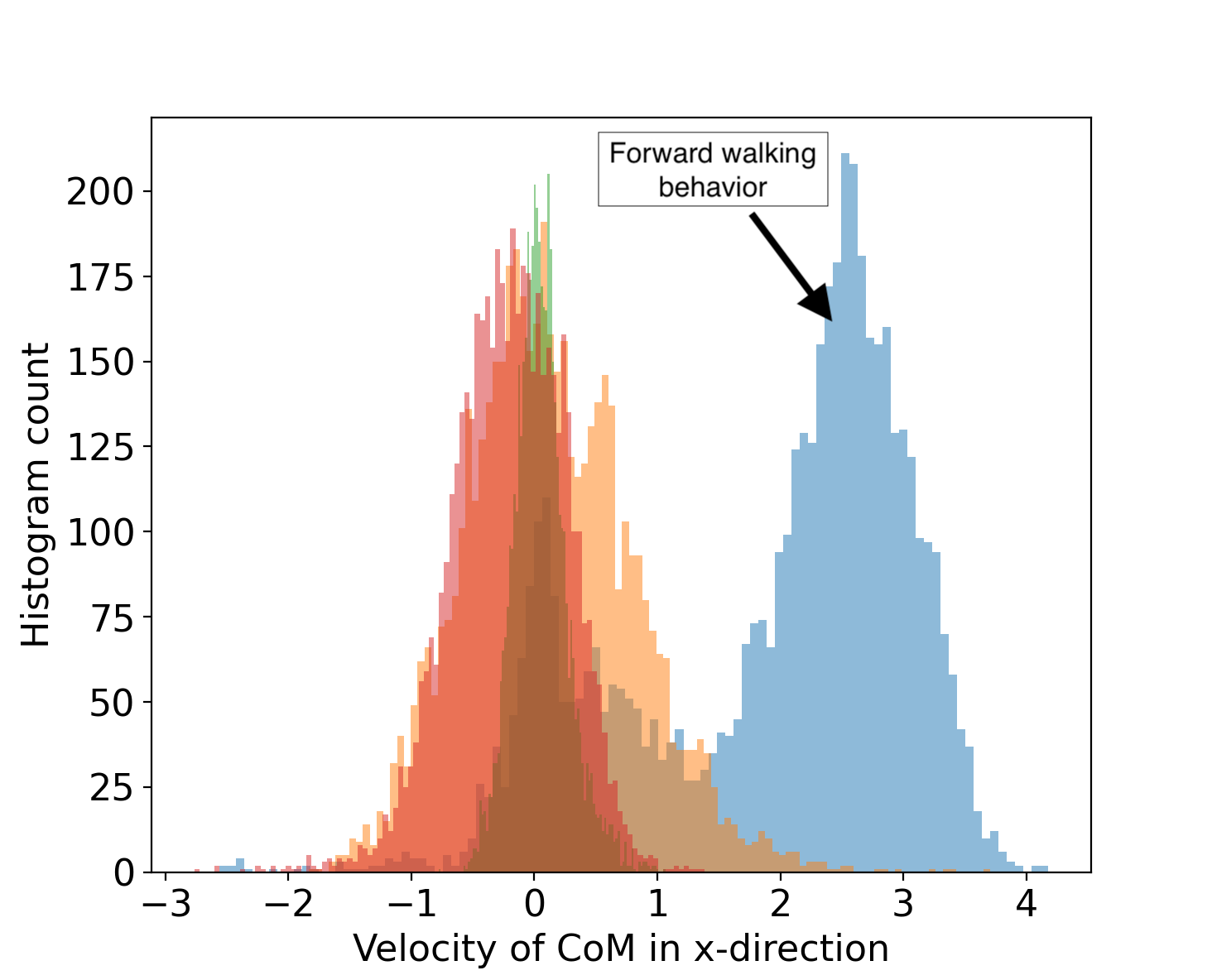}
        \caption{}
        \label{fig:walker_hist}
    \end{subfigure}\hfill
    \caption{(a)-(b) Rendering of the Hopper and Walker tasks, respectively; (c)-(d) Center-of-mass velocity histograms for Hopper and Walker, respectively, when trained with QD-{\em GenDICE}-JS. Arrows point to the emergence of locomotion in one of the policies from the ensemble.}
    \label{fig:walk_hop_overall}
\end{figure}

\textbf{Diversity helps in the absence of environmental rewards.} Designing a task-relevant reward function is typically laborious and error-prone. In the absence of an external reward signal, the diversity objective alone has been previously demonstrated to lead to useful skills~\citep{eysenbach2018diversity}. We test the efficacy of our method in this setting using the Hopper and Walker tasks from Gym (Figures~\ref{fig:hopper_env}-~\ref{fig:walker_env}) but modify the code to return a zero reward for each timestep. Thus, the gradient from the quality-enforcing component in Equation~\ref{eq:svpg} is absent and the QD ensemble is trained only to maximize diversity. After training, we generate a few trajectories with the constituent policies and plot histograms with the velocity of the center-of-mass of the bot on the $x$-axis and the respective counts on the $y$-axis (Figures~\ref{fig:hopper_hist}-~\ref{fig:walker_hist}). Both tasks are learned with QD-{\em GenDICE}-JS and each policy is colored differently. We note that the hopping (respectively walking) behavior {\em emerges} even in the absence of Gym rewards, suggesting that diversity is a strong signal for learning interesting skills.

\subsection{Hyperparameters}\label{appn:hyperparameters}
\begin{table}[H]
\centering
\setlength\belowcaptionskip{-15pt}
\resizebox{0.6\textwidth}{!}{%
\begin{tabular}{c|c}
\multicolumn{1}{c|}{Hyper-parameter}         & \multicolumn{1}{c}{Value}    \\ \hline \hline        

\multicolumn{1}{c|}{Kernel temperatures}             & \multicolumn{1}{c}{$k_\text{JS} (0.5)$, $k_\text{KLS} (1.0)$}   \\ \hline   

\multicolumn{1}{c|}{{\em GenDICE} penalty coefficient}             & \multicolumn{1}{c}{10.}   \\ \hline       

\multicolumn{1}{c|}{Policy network} & \multicolumn{1}{c}{2 hidden layers, 64 hidden dim, tanh}   \\ \hline

\multicolumn{1}{c|}{RL algorithm}             & \multicolumn{1}{c}{PPO (clip=0.2), lr=1e-4}   \\ \hline                                              
\multicolumn{1}{c|}{$\gamma \text{ {\small(Discount)}}, \lambda$ \text{{\small(GAE~\citep{schulman2015high})}}}             & \multicolumn{1}{c}{0.99, 0.95}   \\ \hline   

\end{tabular}}
\end{table}

The networks trained in each of the distribution ratio estimation methods (Section~\ref{subsec:estimatingRatio}) are as follows:
\begin{itemize}
    \item \textbf{{\em NCE}:} For each policy $\pi_i$, an estimator for its stationary distribution $\rho_i$. These are \{2 hidden layers, 100 hidden dim, tanh\} networks.
    \item \textbf{{\em DualDICE}:} For each $\zeta_{ij}$ estimation, a network for the $\nu$ function and a network for the $g$ function (Appendix~\ref{appn:dualdice}). These are \{2 hidden layers, 100 hidden dim, tanh\} networks.
    \item \textbf{{\em ValueDICE}:} For each $\zeta_{ij}$ estimation, a network for the $\nu$ function. These are \{2 hidden layers, 100 hidden dim, tanh\} networks.
    \item \textbf{{\em GenDICE}:} For each $\zeta_{ij}$ estimation, a network for the $\zeta_\theta$ function and a network for the $g$ function (Appendix~\ref{appn:gendice}). These are \{2 hidden layers, 100 hidden dim, tanh\} networks.
\end{itemize}

\end{document}